%% file: egpaper_final.tex
\documentclass[10pt,twocolumn,letterpaper]{article}

\usepackage{booktabs}
\usepackage{cvpr}
\usepackage{times}
\usepackage{epsfig}
\usepackage{graphicx}
\usepackage{amsmath}
\usepackage{amssymb}
\usepackage{balance}
\makeatletter
\g@addto@macro\normalsize{%
  \setlength\abovedisplayskip{1pt}
  \setlength\belowdisplayskip{1pt}
  \setlength\abovedisplayshortskip{1pt}
  \setlength\belowdisplayshortskip{1pt}
}
\makeatother

\usepackage[T1]{fontenc}
\usepackage[utf8]{inputenc}
\usepackage{authblk}
\makeatletter
\renewcommand\AB@affilsepx{ \protect\Affilfont}
\makeatother
\usepackage[breaklinks=true,bookmarks=false]{hyperref}

\cvprfinalcopy 

\def\cvprPaperID{92} 

\ifcvprfinal\fi
\begin{document}

\title{Every Smile is Unique: Landmark-Guided Diverse Smile Generation\vspace{-8mm}}
\author[1,4]{Wei Wang}
\author[2]{Xavier Alameda-Pineda}
\author[1]{Dan Xu}
\author[4]{Pascal Fua}
\author[1,3]{Elisa Ricci}
\author[1]{Nicu Sebe}
\affil[1]{Department of Information Engineering and Computer Science (DISI), University of Trento, Italy, \protect \\}
\affil[2]{Inria, Grenoble-Alpes University, France,}
\affil[3]{Fondazione Bruno Kessler (FBK), Italy.\protect\\}
\affil[4]{Computer Vision Laboratory, EPFL, Lausanne, Switzerland.\protect\\}
\affil[ ]{\tt\small {\{wei.wang,dan.xu,niculae.sebe\}@unitn.it\protect\\} {eliricci@fbk.eu}
{xavier.alameda-pineda@inria.fr}
{pascal.fua@epfl.ch\vspace{-5mm}} 
}

\maketitle

\begin{abstract}
Each smile is unique: one person surely smiles in different ways (\eg closing/opening the eyes or mouth). Given one input image of a neutral face, can we generate multiple smile videos with distinctive characteristics? To tackle this \textbf{one-to-many video generation} problem, we propose a novel deep learning architecture named Conditional Multi-Mode Network (CMM-Net). To better encode the dynamics of facial expressions, CMM-Net explicitly exploits facial landmarks for generating smile sequences. Specifically, a variational auto-encoder is used to learn a facial landmark embedding. This single embedding is then exploited by a conditional recurrent network which generates a landmark embedding sequence conditioned on a specific expression (\eg spontaneous smile). Next, the generated landmark embeddings are fed into a multi-mode recurrent landmark generator, producing a set of landmark 
sequences still associated to the given smile class but clearly distinct from each other. Finally, these landmark sequences are translated into face videos. Our experimental results demonstrate the effectiveness of our CMM-Net in generating realistic videos of multiple smile expressions.
\end{abstract}
\vspace{-5mm}
\section{Introduction}
Facial expressions are one of the --if not \textit{the}-- most prominent non-verbal signals for human communication~\cite{vinciarelli2009social}. For a few decades, researchers in computer vision studied how to automatically recognize such signals~\cite{zen2016learning,Park:2015:SFE:2733373.2806362,Gong:2009:AFE:1631272.1631358,Zhang:2016:MFE:2964284.2967240}. Classically, the analysis of facial expressions has been tackled with a plethora of discriminative approaches, aiming to learn the boundaries between various categories in different video sequence representation spaces. Naturally, these approaches focus on recognizing the dynamics of the different facial expressions. Even if their performance is, specially lately, very impressive, these methods do not posses the ability to reproduce the dynamics of the patterns they accurately classify. How to generate realistic facial expressions is a scientific challenge yet to be soundly addressed.

\begin{figure}[t]
\centering
\includegraphics[width=.92\linewidth]{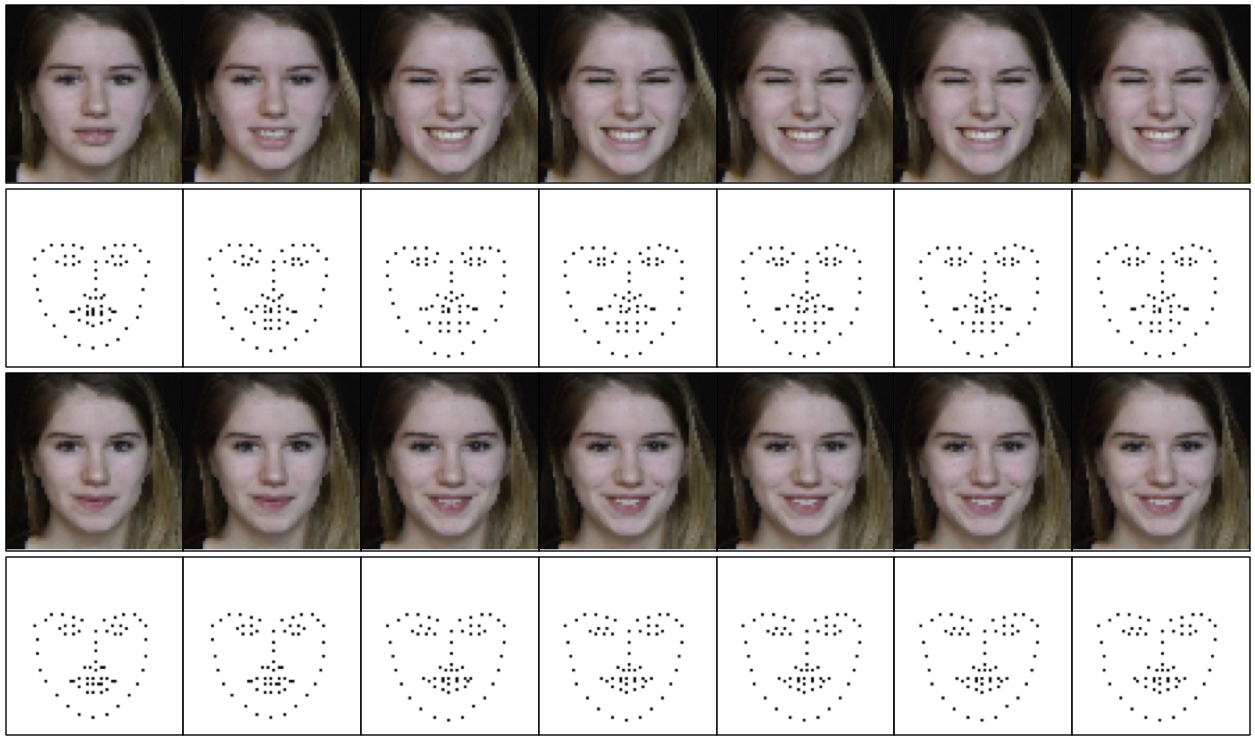}
\caption{Two different sequences of spontaneous smiles and associated landmarks. While there is a common average pattern, 
the changes from one sequence to another are clearly visible.\vspace{-6mm}}
\label{fig:teaser}
\end{figure}


In particular, we are interested in generating \textit{different} facial expressions, for instance, posed \textit{vs}. spontaneous smiles. In reality, one can smile in different ways. As shown in Figure~\ref{fig:teaser}, both videos are spontaneous smile of the same person, but they are quite different (\eg closed vs. open eyes and mouth). The underlying research question is, given one single neutral face, \textit{can we generate diverse face expression videos conditioned on one facial expression label?} 


Thanks to the proliferation of deep neural architectures, and in particular of generative adversarial networks (GAN)~\cite{goodfellow2014generative,denton2015deep} and variational auto-encoders (VAE)~\cite{kulkarni2015deep}, the popularity of image generation techniques has increased in the recent past. Roughly speaking, these methods are able to generate realistic images from encoded representations that are learned in an automatic fashion. Remarkably, the literature on video generation is far less populated 
and few studies addressing the generation of videos~\cite{oh2015action,srivastava2015unsupervised,tulyakov2017mocogan} or the generation of predicted actions in videos~\cite{koppula2016anticipating} exist. 
In this context, it is still unclear how to generate distinct video sequences given a single input image. 

The dynamics of facial expressions, and of many other facial (static and dynamic) attributes are encoded in the \textit{facial landmarks}. For instance, it has been shown that landmarks can be used to detect whether a person is smiling spontaneously or in a posed manner~\cite{dibekliouglu2012you}. Action units (e.g. check raiser, upper lip raiser) are also closely related to both facial expressions and facial landmarks~\cite{king1976luminance}. Therefore, we adopt facial landmarks as a compact representation of the facial dynamics and a good starting point towards our aim. Figure~\ref{fig:teaser} shows an example to further motivate the use of landmarks and to illustrate the difficulty of the targeted problem. Indeed, in this figure we can see two examples of spontaneous smiles and their associated landmarks. The differences are small but clear (e.g., closed vs. open eyes). Therefore, it is insufficient to learn an ``average'' spontaneous smiling sequence. We are challenged with the task of learning distinct landmark patterns belonging to the same class. Thus, given a neutral face, the generation of diverse facial expression sequences of a certain class is a \textit{one-to-many} problem. 



A technology able to generate different facial expressions of the same class would have a positive impact in different fields. For instance, the face verification and facial expression recognition systems would be more robust to noise and outliers, since there would be more data available for training. 
In addition, systems based on artificial 
agents, impersonated by an avatar, would clearly benefit from an expression generation framework able to synthesize distinct image sequences of the same class. Such agents would be able to smile in different ways, as humans do.


In this paper, we propose a novel approach for generating videos of smiling people given an initial image of a neutral face. 
Specifically, we 
introduce a methodological framework which generates various image sequences (i) that correspond to the desired class of expressions (\ie posed/spontaneous smile), (ii) that look realistic and implicitly preserve the identity of the input image and (iii) that have clearly visible differences between them. 
As previously explained, we exploit facial landmarks since they encode the dynamics of facial expressions in an effective manner. 
First, a compact representation of the landmark manifold is learned by means of a variational auto-encoder. 
This representation is further used to learn a conditional 
recurrent network (LSTM) which takes as input the landmarks automatically extracted from the initial neutral face and generates a sequence of landmark embeddings conditioned on a given facial expression. This sequence 
is then fed to a 
multi-mode recurrent landmark generator, which consists of multiple LSTMs and is able to output a set of clearly distinct landmark embedding sequences. Remarkably, the second generating layer does not require additional ground truth to be trained. 
The input face image is then used 
for translating the generated landmark embedding sequences into distinct face videos.
The joint architecture is named Conditional Multi-Mode (CMM) recurrent network. 
We evaluate the proposed method on three public datasets: the UvA-NEMO Smile~\cite{dibekliouglu2012you}, the DISFA~\cite{mavadati2013disfa} and DISFA+~\cite{mavadati2016extended}.

\vspace{-2mm}
\input{related}


\vspace{-1mm}
\section{Conditional Multi-Mode Generation}
\vspace{-1mm}
\label{sec:method}
\begin{figure}[t]
\includegraphics[width=\linewidth]{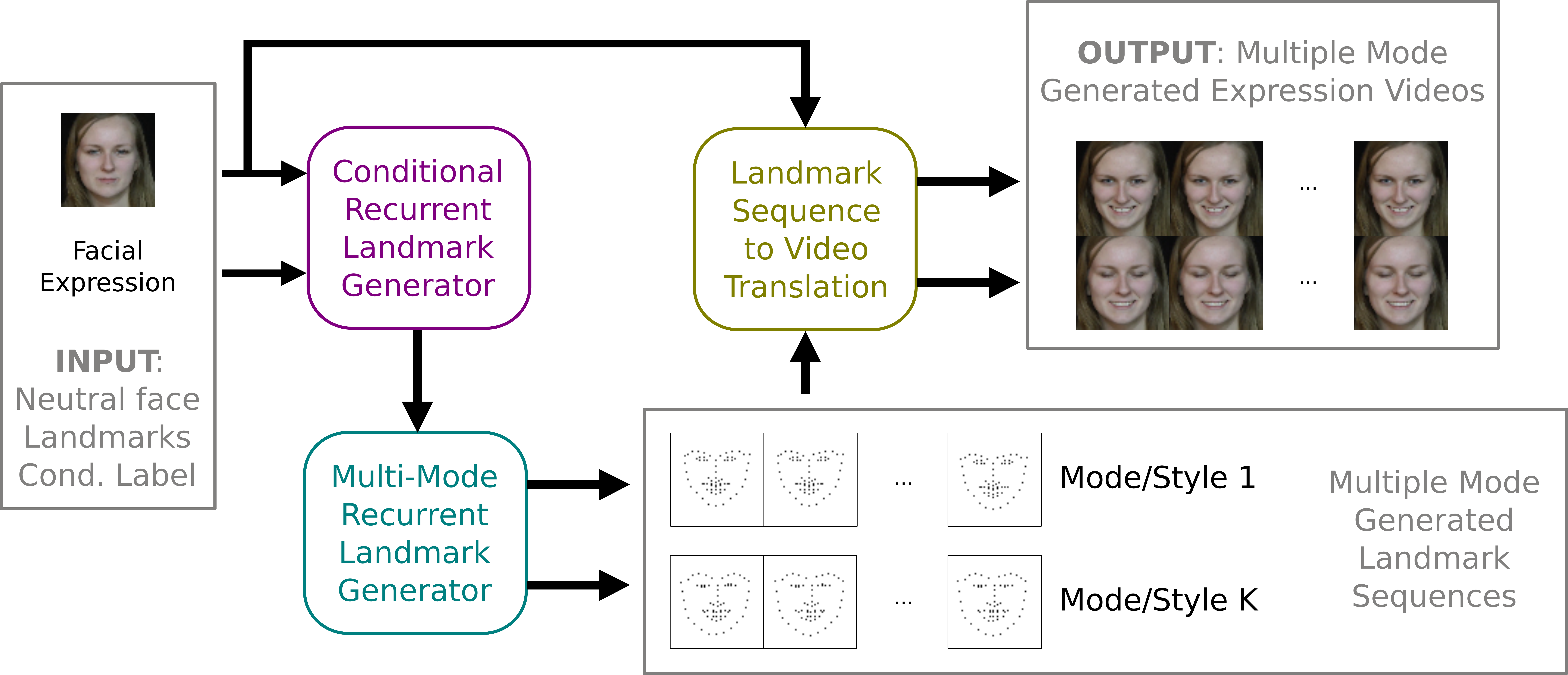}
\caption{{Overview of the proposed framework. The input 
image is used together with the conditioning label to generate a set of $K$ distinct landmark sequences. These landmark sequences guide the neutral face image to translate into face videos.}\vspace{-5mm}}
\label{fig: overview}
\end{figure}
\subsection{Overview}
\vspace{-1mm}
The overall architecture consists of three blocks (see Fig.~\ref{fig: overview}) that are able to generate multiple facial expression sequences corresponding to a person and of a given facial expression class (\eg spontaneous vs. posed smile). First, the conditional recurrent landmark generator (purple box) computes a landmark image from the input face, encodes it into a compact representation and generates a landmark sequence corresponding to the desired facial expression class. Second the multi-mode recurrent landmark generator (turquoise box) receives this sequence and generates $K$ sequences of the same class with clearly distinct features. Finally, the landmark sequence to video translation module (ocher box) receives these landmark sequences and the initial neutral face image to produce the output facial expression videos. The entire architecture is named Conditional Multi-Mode recurrent network. In summary, the input consists of (i) a neutral face image 
and (ii) the desired expression label. The output is a set of $K$ face videos each one containing a different facial expression sequence corresponding to the specified class. In the following we describe the three main blocks in details.
\vspace{-1mm}
\subsection{Conditional Recurrent Landmark Generator}
\vspace{-1mm}
The conditional recurrent landmark generator (magenta box in Figure~\ref{fig: method_detail}) receives a face image and a conditioning facial expression label as inputs. We automatically extract the landmark image from the face image and encode it using a standard VAE~\cite{kulkarni2015deep} into a compact embedding, denoted as $h_0$. Details are in Section~\ref{sec:exp}.
A conditional Long-Short Term Memory (LSTM) recurrent neural network is used to generate a sequence of $T$ facial landmark embeddings, denoted by $\mathbf{h}=(h_1,\ldots,h_T)$. The conditional label is encoded and input at all time steps of the conditional LSTM.
The embedding sequence $\mathbf{h}$ is further decoded into a landmark image sequence, $\mathbf{x}=(x_1,\ldots,x_T)$, which is encouraged to be close to the training landmark image sequence $\mathbf{y}$ by computing a pixel-wise binary cross-entropy (BCE) loss. In more detail, given a training set of $N$ sequences of length $T$, $\{\mathbf{y}^n=(y^{n}_{1},\ldots,y^{n}_{T})\}_{n=1}^{N}$, the loss of the conditional recurrent landmark generator writes:
\begin{equation}
{\cal L}_\textrm{BCE} = \sum_{n,t=1}^{N,T} y^{n}_{t}\odot\log x^{n}_{t} + (1-y^{n}_{t})\odot\log (1-x^{n}_{t}),
\end{equation}
where $\odot$ and $\log$ denote the element-wise product and natural logarithm operations respectively.\footnote{To keep the notation simple, the addition over the pixels in the image is not explicit. In addition, the upper index denotes correspondence to the $n$-th training sample.}

If one needs to generate face videos of a given class this methodological apparatus would suffice. However, how could we generate diverse sequences of the same class given one single image? 
First, this would require recording several times the ``same'' facial expression of a person with different patterns, which is particularly difficult for spontaneous facial expressions. Even if such dataset was ready, it is still unclear how to make one single conditional LSTM to generate diverse distinct sequences: a straightforward training would do nothing else but learn the average landmark sequence. The module described in the next section is specifically designed to overcome this limitation.





\vspace{-1mm}
\subsection{Multi-Mode Recurrent Landmark Generator}
\vspace{-1mm}
\begin{figure*}[!htb]
\centering
\includegraphics[width=.95\linewidth]{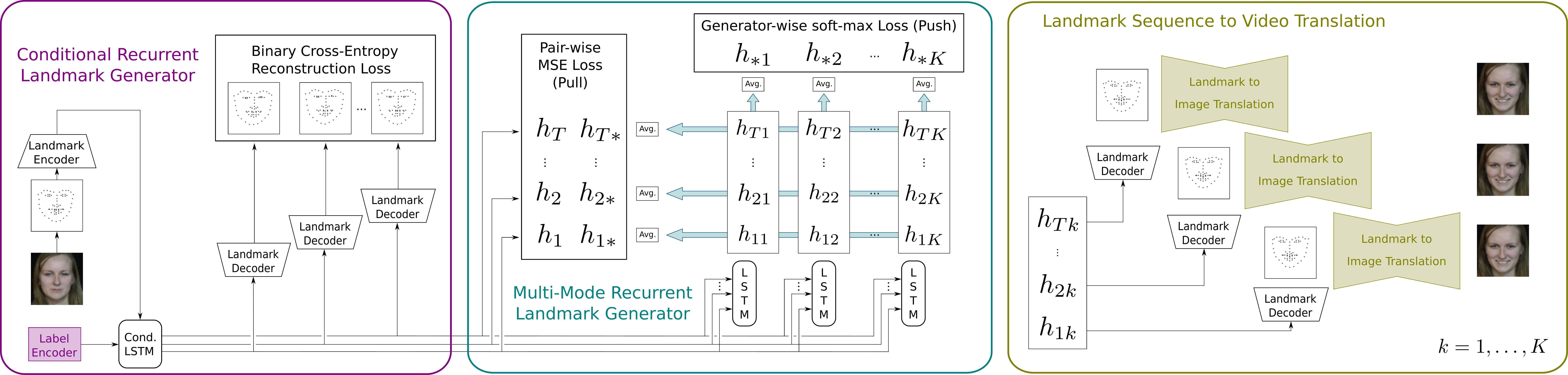}
\caption{Detail of the conditional multi-mode recurrent network. The left block (magenta) encodes the landmark image and generates a sequence of landmark embeddings according to the conditioning label. The second block (turquoise) generates $K$ different landmark embedding sequences. Finally, the third block (ocher) translates each of the sequences into a face video.\vspace{-5mm}}
\label{fig: method_detail}
\end{figure*}

As briefly discussed in the previous section, we would like to avoid recording several sequences of the same person, since it may be a tedious process and, more importantly, spontaneous facial expressions are scarce and hard to capture. Ideally, the network module used to generate multiple modes should not require more supervision than the one already needed by the previous module.

We designed the multi-mode recurrent landmark generator (turquoise box of Fig.~\ref{fig: method_detail}) on these grounds. It consists of $K$ LSTMs, whose input is the sequence of embeddings generated by the conditional LSTM: $h_1,\ldots,h_T$ and the output is a set of $K$ generated sequences $\{\mathbf{h}_k=(h_{1k},\ldots,h_{Tk})\}_{k=1}^K$. In a nutshell this is a one-to-many sequence mapping that has to be learned in an unsupervised fashion. On the one side, we would like the sequences to exhibit clearly distinct features. On the other side, the sequences must encode the desired facial expression. Intuitively, the method finds an optimal trade-off between \textit{pushing} the sequences to be distinct and \textit{pulling} them towards a common pattern. While the differentiating characteristics can happen at various instants in time, the common pattern must respect the dynamics of the smile. This is why, as formalized in the following, the pushing happens over the temporally-averaged sequences while the pulling is encourage on the mode/generator-wise averages.

Formally, we define $(h_{1*},\ldots,h_{T*})$ as the sequence of mode-wise averaged generated landmark encodings (horizontal turquoise arrows) and $\{h_{*k}\}_{k=1}^K$ as the set of temporally-averaged landmark embedding sequences. With this notation, and following the intuition described in the previous paragraph the push-pull loss is defined as follows. First, we impose a mean squared error loss between the generator-wise average $(h_{1*},\ldots,h_{T*})$ and the sequence generated by conditional LSTM $(h_1,\ldots,h_T)$:
\begin{equation}
{\cal L}_{\textrm{Pull}} = \sum_{n,t=1}^{N,T} \|h_t^n - h_{t*}^n\|_2.
\end{equation}

Second, inspired by the multi-agent diverse GAN~\cite{ghosh2017multi}, we use the cross-entropy loss so as to discriminate between the sequences obtained from the $K$ generators:
\begin{equation}
{\cal L}_{\textrm{Push}} = {-}\sum_{n,k=1}^{N,K} \log \phi_k(h_{*k}^n),
\end{equation}
where $\phi_k$ represents the $k$-th output of the discriminator (a fully connected layer followed by a soft-max layer). Therefore, the overall architecture is GAN-flavored in the sense that the hierarchical LSTMs are topped with a discriminator to differentiate between the various generators. Importantly, this discriminative loss is complementary with the BCE. The entire loss pushes the multiple sequences far away from each other while encouraging the overall system to behave accordingly to the training data. In GAN, the generator and discriminator compete with each other. In contrast, they work cooperatively in our module.


Note that the combination of the conditional and multi-mode landmark recurrent generators has several advantages. First, as already discussed, the multi-mode generator does not require more ground truth than the conditional one. Second, thanks to the push-pull loss, the generated sequences are pushed to be diverse while pulled to stay around a common pattern. Third, while the conditional block is, by definition, conditioned by the label, the second block is transparent to the input label. This is important on one hand because we do not have a specific multi-mode recurrent landmark generator per conditional label, thus reducing the number of network parameters and the amount of data needed for training. On the other hand, because by training the multi-mode generator with data associated to different class labels, it will focus on facial attributes that are not closely correlated with the conditioning labels, and one can expect a certain generalization ability when a new facial expression is added in the system.

\subsection{Landmark Sequence to Video Translation}
The last module of the architecture is responsible for generating the face videos, \ie, translating the facial landmark embeddings generated by the two first modules into image sequences. To do so we employ a U-Net like structure~\cite{isola2016image} after the facial landmark image decoder. Let $z^n_0$ denote the input neutral face image associated to the $n$-th training sequence. Together with the facial landmark images $\{\mathbf{y}^n=(y^{n}_{1},\ldots,y^{n}_{T})\}_{n=1}^{N}$ already used to train the previous modules, the dataset contains the face images (from which the facial landmarks are annotated) denoted by $\{\mathbf{z}^n =(z^{n}_{1},\ldots,z^{n}_{T})\}_{n=1}^{N}$.

In order to train the translation module we employ a combination of a reconstruction loss and an adversarial loss, since we want the generated images to be \textit{locally} close to the ground-truth  and to be \textit{globally} realistic. Let $w^n_t(\theta_{\cal G}) = {\cal G} (y^n_t,z^n_0;\theta_{\cal G})$ denote the face image generated with the facial landmark image $y^n_t$ and the neutral face image $z^n_0$, with parameters $\theta_{\cal G}$. The reconstruction loss writes:
\begin{equation}
{\cal L}_{\textrm{Rec}} = \sum_{n,t=1}^{N,T} \|z^n_t-w^n_t(\theta_{\cal G})\|_1.
\end{equation}

The adversarial loss is defined over real $[z^n_0,z^n_t]$ and generated $[z^n_0,w^n_t]$ image pairs:
\begin{align}
{\cal L}_{\textrm{Adv}} &= \sum_{n,t=1}^{N,T} \log {\cal D}([z^n_0,z^n_t];\theta_{\cal D}) \nonumber\\
& + \sum_{n,t=1}^{N,T} \log (1-{\cal D}([z^n_0,w^n_t(\theta_{\cal G})];\theta_{\cal D})) \label{eq:adv_loss}
\end{align}

When the generator is fixed, the discriminator is trained to maximize~(\ref{eq:adv_loss}). When the discriminator is fixed, the generator is trained to jointly minimize the adversarial and reconstruction losses with respect to $\theta_{\cal G}$:
\begin{equation}
\sum_{n,t=1}^{N,T} \|z^n_t-w^n_t(\theta_{\cal G})\|_1 + \log (1-{\cal D}([z^n_0,w^n_t(\theta_{\cal G})];\theta_{\cal D}))
\end{equation}

Furthermore, inspired by~\cite{yoo2016pixel}, we use the adversarial loss at the pixel-level of the feature map. In other words, there is one label per pixel of the coarsest feature map, instead of one label per image. Intuitively, this loss should be able to focus in many parts of the image individually, instead of seeing the image as a whole.



\subsection{Training Strategy}
The training of the CMM architecture is done in three phases. First, we train the landmark embedding VAE so as to reconstruct a set of landmark images $\{y^n_{t}\}_{n,t=1}^{N,T}$. This VAE is trained for 50 epochs before the conditional LSTM is added. The second phase consists on fine-tuning the VAE and training the first layer LSTM on the dataset of sequences of landmark images $\{\mathbf{y}^n\}_{n=1}^N$ for 20 epochs. The third stage consists on adding the multi-model recurrent landmark generator. Therefore the VAE and LSTM are fine tuned at the same time the $K$ different LSTMs are learned from scratch. This phase includes the reconstruction, pull and push loss functions previously defined and lasts 10 epochs. Finally, the landmark sequence to video translation module is trained apart from the rest for 20 epochs. More details can be found in the supplementary material. 

\vspace{-2mm}
\section{Experimental Validation}
\vspace{-1mm}
\label{sec:exp}
\subsection{Experimental Setup}
\vspace{-1mm}
\paragraph{Datasets and Preprocessing.}
We demonstrate the effectiveness of the proposed approach by performing experiments on three publicly available datasets, namely: UvA-NEMO Smile~\cite{dibekliouglu2012you}, DISFA~\cite{mavadati2013disfa} and DISFA+~\cite{mavadati2016extended}. 

The UvA-NEMO dataset~\cite{dibekliouglu2012you} contains 1240 videos, 643 corresponding to posed smiles and 597 to spontaneous ones. The dataset comprises 400 subjects (215 male and 185 female) with different ages ranging form 8 to 76 (50 subjects wear glasses). The videos are sampled at 50 FPS and frames have a resolution of $1920 {\times} 1080$ pixels, with an average duration of 3.9~s. The beginning and the end of each video corresponds to a neutral expression. 

\begin{figure}[t]
\includegraphics[width=\linewidth]{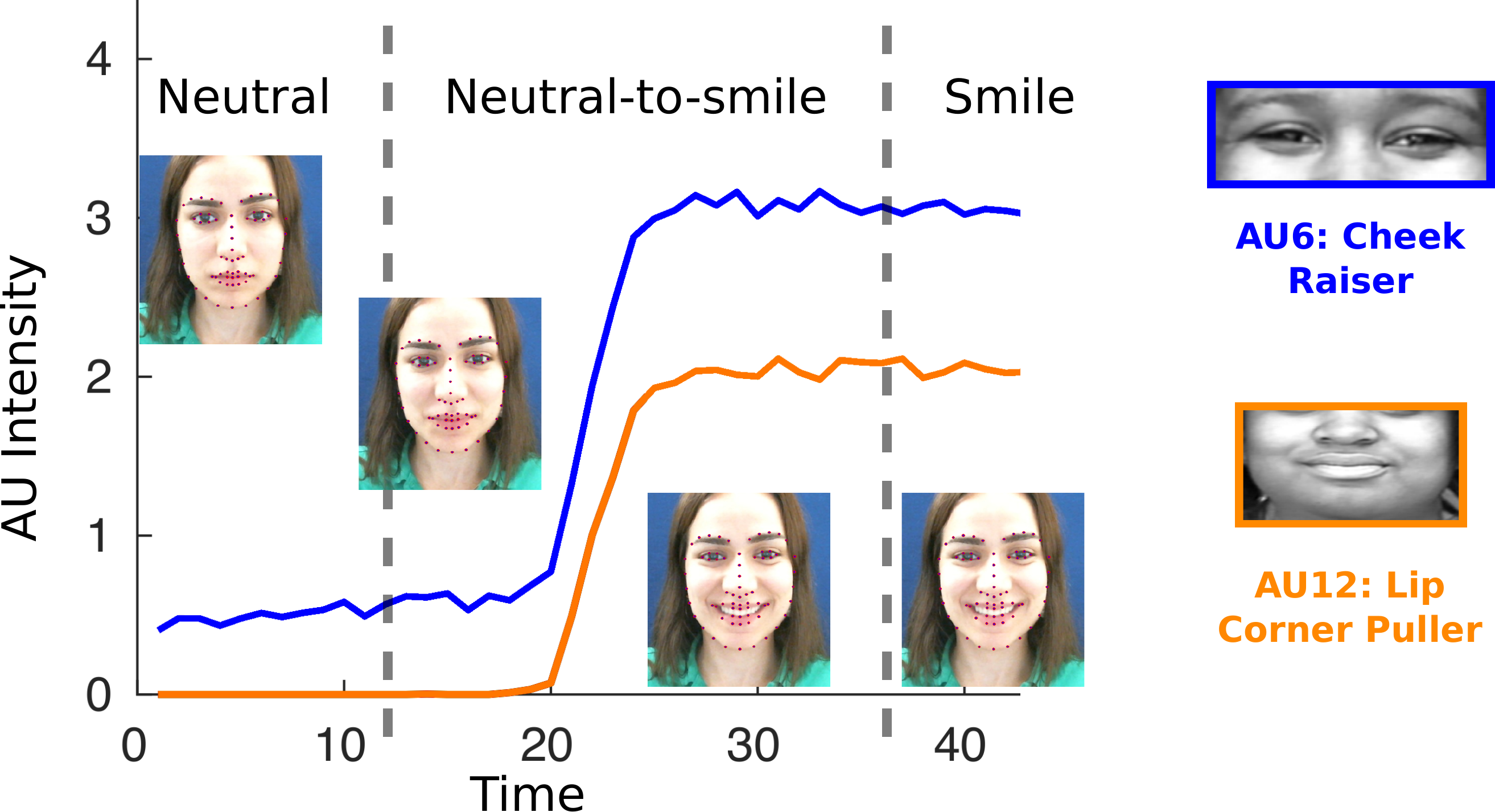}
\caption{Action unit dynamics in neutral-to-smile transitions: \textit{cheek raiser} and \textit{lip corner puller}.\vspace{-5mm}}
\label{fig: action unit}
\vspace{-2mm}
\end{figure}

The DISFA dataset~\cite{mavadati2013disfa} contains videos with spontaneous facial expressions. In the dataset there are 27 adult subjects (12 females and 15 males) with different ethnicities. The videos are recorded at 20 FPS and the resolution is $1024{\times}768$ pixels. While the dataset contains several facial expressions, in this work we only consider smile sequences and manually segmented the videos to isolate spontaneous smiles, obtaining 17~videos in total.
To gather the associated posed smiles, we also consider the DISFA+ dataset~\cite{mavadati2016extended} which contains posed smile expression sequences for nine individuals present in the DISFA dataset. 


\begin{figure*}[t]
\centering
\includegraphics[width=.95\linewidth]{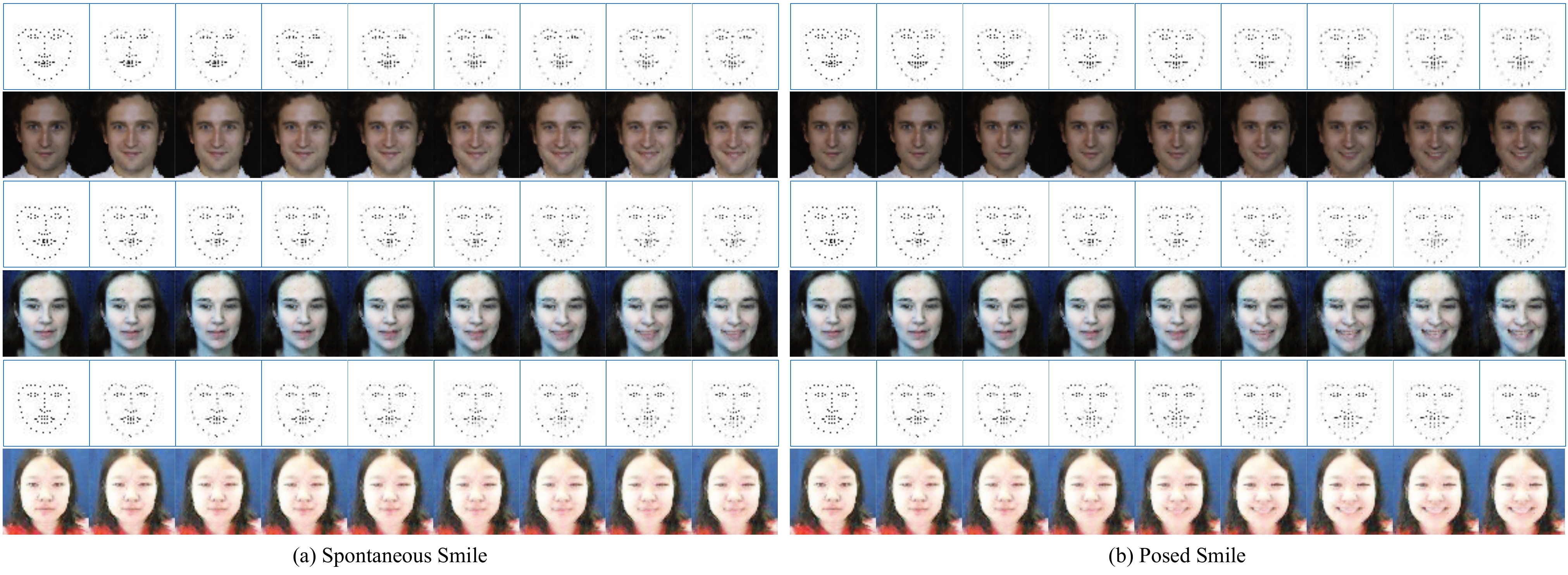}
\vspace{-2mm}
\caption{Landmark sequences generated with the first block of our CMM-Net. The associated face images are obtained using the landmark sequence to video translation block. The left block corresponds to generated spontaneous smiles, while the right block to posed smiles. The three row pairs correspond to the UvA-NEMO, DISFA \& DISFA+ datasets respectively. Images better seen at magnification.}
\vspace{-3mm}
\label{fig:conditioning}
\end{figure*}

\begin{figure*}[t]
\centering
\includegraphics[width=.95\linewidth]{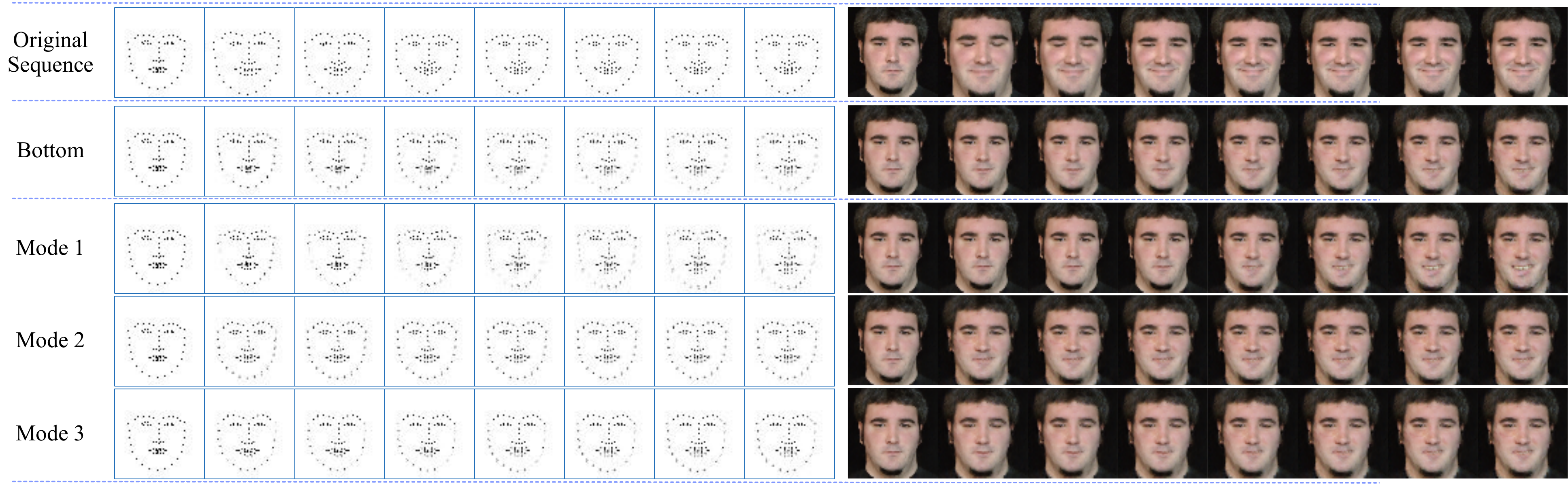}
\vspace{-1mm}
\caption{Multi-mode generation example with a sequence of the UvA-NEMO dataset: landmarks (left) and associated face images (right). The rows correspond to the original sequence, output of the Conditional LSTM, and output of the Multi-Mode LSTM (last three rows).\vspace{-5mm}}
\label{fig:lmk_sequence}
\end{figure*}

The proposed CMM-Net framework requires training sequences of both posed and spontaneous smiles, as well as the associated landmarks. To collect the training data we process the video sequences from the original datasets and extract the subsequences associated to smile patterns. To do that, we rely on action Units (AUs)~\cite{tian2001recognizing} and specifically on the \textit{cheek raiser} and \textit{lip corner puller} AUs. Indeed, we extract the intensity variations of these two AUs with the method in~\cite{baltruvsaitis2015cross}. As shown in Figure~\ref{fig: action unit}, the intensity variations of these two action units are very characteristic of neutral-to-smile (N2S) sequences.
Similar to the pre-processing steps of other works~\cite{wang2018recurrentface,wang2016recurrentface}, we also perform face alignment on the extracted sequences using OpenFace~\cite{baltruvsaitis2016openface}, aligning the faces with respect to the center of the two eyes horizontally and to the vertical line passing through the center of the two eyes and the mouth. 
We notice that in these datasets, the average N2S length is $T=32$ frames, with tiny variations. We sample $T$ frames from the first phase of each video. If the number of video frames in the N2S phase is less than $T$, we pad the sequence with subsequent frames. Images are resized to $64\times64$ pixels. The facial landmarks are extracted using~\cite{baltrusaitis2013constrained}, and $64\times64$ binary images are created from them.
In case of the UvA-NEMO dataset we follow the splitting protocol of~\cite{dibekliouglu2012you} and use 9 splits for training and the 10-th for test. For the paired DISFA-DISFA+ sequences, we randomly select two thirds of the videos for training and the rest for testing. \vspace{-3mm}

\vspace{-2mm}
\paragraph{Network Architecture Details.} The face-image to landmark-image VAE consists of a symmetric convolutional structure with five layers. The first four layers are Conv(4,2,1) (kernel size, stride and padding) with 64, 128, 256 and 512 output channels respectively. All of them have a Leaky ReLU layer and, except for the first one, they use batch normalization. The final layer models the mean and standard deviation of the VAE and are two Conv(4,1,0) layers with 100 output channels each. After the sampling layer, there are the symmetric five convolutional layers with the same parameters as the encoder and 512, 256, 128, 64, and 1 output channels. While the first four layers have a Leaky ReLU layer and use batch normalization, the last layer's output is a sigmoid.

\begin{table*}[t]
\centering
\caption{Quantitative Analysis. The SSIM and Inception Score.\vspace{1mm}}
\label{tab: SSIM_IS_score}
\setlength\tabcolsep{5.5pt}
\resizebox{0.9\textwidth}{!}{%
\begin{tabular}{ccccccccccccccccc}
\toprule
 && \multicolumn{3}{c}{UvA-NEMO Spont.} && \multicolumn{3}{c}{UvA-NEMO Posed} && \multicolumn{3}{c}{DISFA Spont.} && \multicolumn{3}{c}{DISFA+ Posed} \\ \midrule
Model            && IS          & $\Delta$IS        & SSIM        && IS        & $\Delta$IS      & SSIM      && IS           & $\Delta$IS        & SSIM        && IS         & $\Delta$IS       & SSIM      \\ \midrule
Original     && 1.419       & -           & -           && 1.437     & -         & -         && 1.426        & -           & -           && 1.595      & -          & -         \\ 
Video GAN        && 1.576       & 0.157       & 0.466       && 1.499     & 0.062     & 0.450     && 1.777        & 0.351       & 0.243       && 1.547      & 0.048      & 0.434     \\ 
CRA-Net && 1.311       & 0.108       & 0.553       && 1.310     & 0.127     & 0.471     && 1.833        & 0.407       & 0.749       && 1.534      & 0.061      & 0.839     \\ 
CMM-Net              && 1.354       & 0.065       & 0.854       && 1.435     & 0.002     & 0.827     && 1.447        & 0.021        & 0.747       && 1.533      & 0.062      & 0.810     \\ \bottomrule
\end{tabular}}
\vspace{-1mm}
\end{table*}

\begin{figure*}[t]
\centering
\vspace{-2mm}
\includegraphics[width=.95\linewidth]{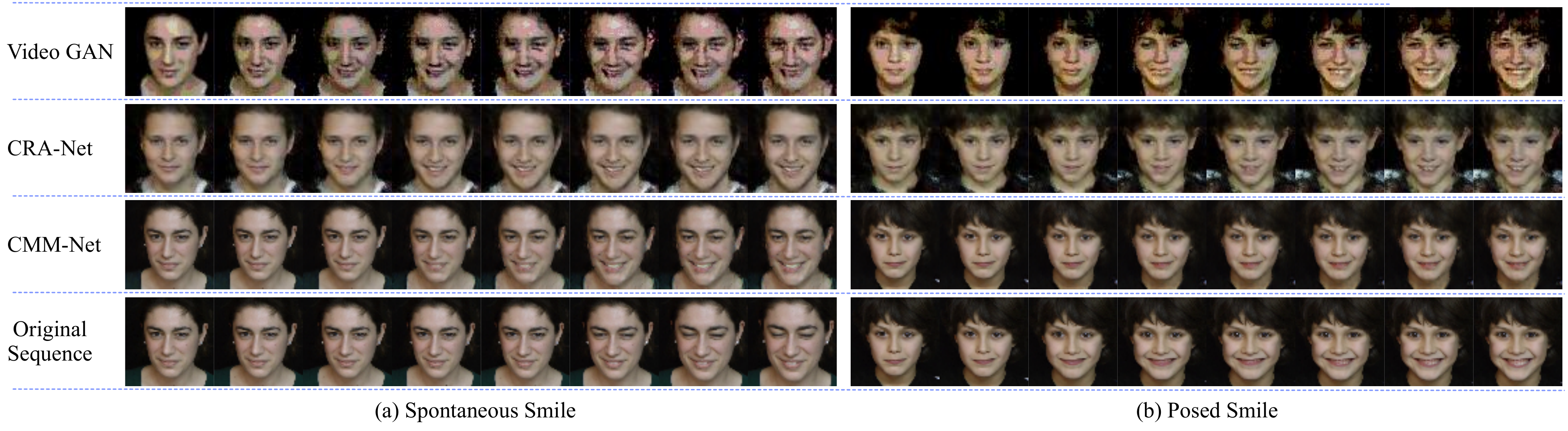}
\vspace{-1mm}
\caption{Qualitative comparison. From top to bottom: original sequence, Video-GAN, CRA-Net and CMM-Net. Video-GAN introduces many artifacts compared to the other two. CRA-Net learn the smile dynamics, but fail to preserve the identity, as opposed to CMM-Net which produces realistic smiling image sequences.\vspace{-5mm}}
\label{fig: comparison_qualitative}
\end{figure*}

The generator of the adversarial translation structure is a fully convolutional auto-encoder network with 6 Conv(4,2,1) layers with 64, 128, 256, 512, 512 and 512 output channels. The first five convolutional layers use a Leaky ReLU, and except for the first, batch normalization. The last layer uses plain ReLU. The decoder has the same structure as the encoder. All layers except the last one use ReLU and batch normalization, and the last one uses a hyperbolic tangent. Notice that the number of input channels is four (neutral face image plus facial landmark image) and the number of output channels is three.


The discriminator of the adversarial translation structure has three Conv(4,2,1) and two Conv(4,1,1) with 64, 128, 256, 512 and 1 output channels respectively. While all except the last one are followed by a Leaky ReLU, only the three in the middle use batch normalization. Recall that, since the input of the discriminator are image pairs, the input number of channels is six.
More details can be found in the supplementary material.

\vspace{-5mm}
\paragraph{Baselines.} The literature on data-driven automatic video generation is very limited and no previous works have considered the problem of smile generation. Therefore, we do not have direct methods to compare with. However, in order to evaluate the proposed approach we compare with the Video-GAN model~\cite{vondrick2016generating}, even if it has not been specifically designed for face videos. Importantly, since one of the motivations of the present study is to demonstrate the importance of using facial landmarks, we also compare to a variant of the proposed approach that learns an embedding from the face images directly, instead from landmark images, and we call it conditional recurrent adversarial network (CRA-Net). The CRA-Net has the same structure as the bottom layer conditional recurrent landmark generator. The difference is that a discriminator is added on the top of the generated images to improve the image quality.



\vspace{-2mm}
\subsection{Qualitative Evaluation} 
\vspace{-2mm}
We first show that the proposed Conditional Recurrent Landmark Generator is able to synthesize landmark sequences corresponding to different conditioning labels. Figure~\ref{fig:conditioning} shows the landmark images obtained for the same neutral face and different conditioning labels (\ie spontaneous/posed). From these results, it is clear that the generated landmarks (and associated face images) follow different dynamics depending on the conditioning label.

To demonstrate the effectiveness of the proposed Multi-Mode Recurrent Landmark Generator block, we also show the results associated to generating multiple landmark sequences with different styles. In this experiment we set $K=3$. Given a neutral face, the associated landmark image and the conditioning label, we can obtain 4 landmark sequences: the first is obtained from the Conditional LSTM, while the others are generated through the $K$ LSTMs corresponding to different styles. An example of the generated landmark sequences for a posed smile is shown in Fig. \ref{fig:lmk_sequence}, together with the associated images recovered using the translation block. Our results show that the landmark sequence generated by the Conditional LSTM is very similar to the original sequence. Moreover, the landmark images corresponding to multiple styles exhibit clearly distinct patterns, \eg the subject smiles with a wide open mouth ($3^\textrm{rd}$ row), with mouth closed ($4^\textrm{th}$ row) and with closed eyes ($5^\textrm{th}$ row).

Figure~\ref{fig: comparison_qualitative} reports generated sequences of different methods, to benchmark them with the proposed CMM-Net. The first row shows results obtained with Video-GAN~\cite{vondrick2016generating}, the second row corresponds to CRA-Net, 
the third row is obtained with the proposed CMM-Net, and the fourth row is the original image sequence. %
From the results, we can observe that the images generated by Video-GAN contain much more artifacts than the other two methods. The images of CRA-Net are quite realistic, meaning that even without learning the landmark manifold space the dynamics of the smile is somehow captured. However, we can clearly see that the identity of the person is not well preserved, and therefore the sequences look unrealistic. The CMM-Net decouples the person identity from the smile dynamics (considering the translation and the recurrent blocks, respectively), and thus being able to generate smooth smiling sequences that preserve the identity of the original face.

\vspace{-2mm}
\subsection{Quantitative Analysis}
\vspace{-2mm}
To further demonstrate the effectiveness of our framework we conduct a quantitative analysis computing some objective measures of the reconstruction quality, performing a user study and measuring the AUs dynamics of generated sequences.
\vspace{-5mm}

\begin{table}[t]
\centering
\caption{CMM-Net vs Video-GAN and CMM-Net vs CRA-Net: percentage (\%) of the preferences of the generated videos.\vspace{1mm}}
\label{tab:user_study}
\scalebox{0.8}{
\begin{tabular}{ccc} \toprule
Models  & Spontaneous Smile		& Posed Smile\\ \midrule
Video-GAN~\cite{vondrick2016generating}&            10.14            &          7.24        \\ 
CMM-Net   &            85.14            &          83.68       \\
$\sim$   		&             4.72            &           9.08    \\ \midrule
CRA-Net&            17.76             &         11.94        \\ 
CMM-Net   &             54.87           &           59.72        \\
$\sim$   		&             27.37            &           28.33     \\ \bottomrule
\end{tabular}}
\vspace{-6mm}
\end{table}

\paragraph{Objective Measures.} Structure similarity (SSIM)~\cite{wang2004image} and inception score (IS)~\cite{salimans2016improved} are employed to measure the quality of the synthesized images. Table~\ref{tab: SSIM_IS_score} reports these two scores for the benchmarked methods. The interpretation of these scores must be done with care. Usually, and specially for SSIM, larger image similarity score corresponds to more realistic images. However, high quality images do not always correspond to large IS scores, as observed in~\cite{ma2017pose,shi2016real}. Indeed, a generative model could collapse towards low-quality images with large inception score. This effect is also observed in our experiments if we put Table~\ref{tab: SSIM_IS_score} and Figure~\ref{fig: comparison_qualitative} side to side. This is why we also report the score difference between the generated sequence and the original sequence as $\Delta$IS. Intuitively, the smaller this difference is, the more similar is the quality of the generated images to the quality of the original images. Overall, CMM-Net have the higher SSIM score and the smallest difference in IS score. 


\vspace{-5mm}

\paragraph{User Study.} To further demonstrate the validity of the proposed framework, we perform a user-study and compare the videos generated by CMM-Net to the ones generated by Video GAN and CRA-Net.
The Video-GAN approach in~\cite{vondrick2016generating} can only 
generate videos given an input frame but does not employ conditioning labels. In order to perform a comparison we train two different models corresponding to the two different smiling labels. 
To compare with each of the baseline, we show a subject a pair of videos (one generated by our CMM-Net and the other by the baseline method) and ask \textit{Which video looks more realistic?}. We prepared 37 video pairs and invited 40 subjects to do the evaluation. We collected 1480 ratings for each of the experiments. 
Table~\ref{tab:user_study} shows the preferences expressed by the annotators ($\%$) both for spontaneous and posed smiles.
The symbol $\sim$ indicates that the two videos are rated as similar. When we compare the CMM-Net with the Video GAN baseline (Table~\ref{tab:user_study}), most annotators prefer the videos generated by our CMM-Net. This is not surprising: by visually inspecting the frames we observe that several artifacts are present in the sequences generated with Video-GAN (see Fig.\ref{fig: comparison_qualitative}). Furthermore, comparing our approach with CRA-Net (Table~\ref{tab:user_study}), we still observe that most annotators prefer images obtained with CMM-Net, confirming the benefit of adopting landmark for face video generation.



\begin{figure}[t]
\centering
\includegraphics[width=\linewidth]{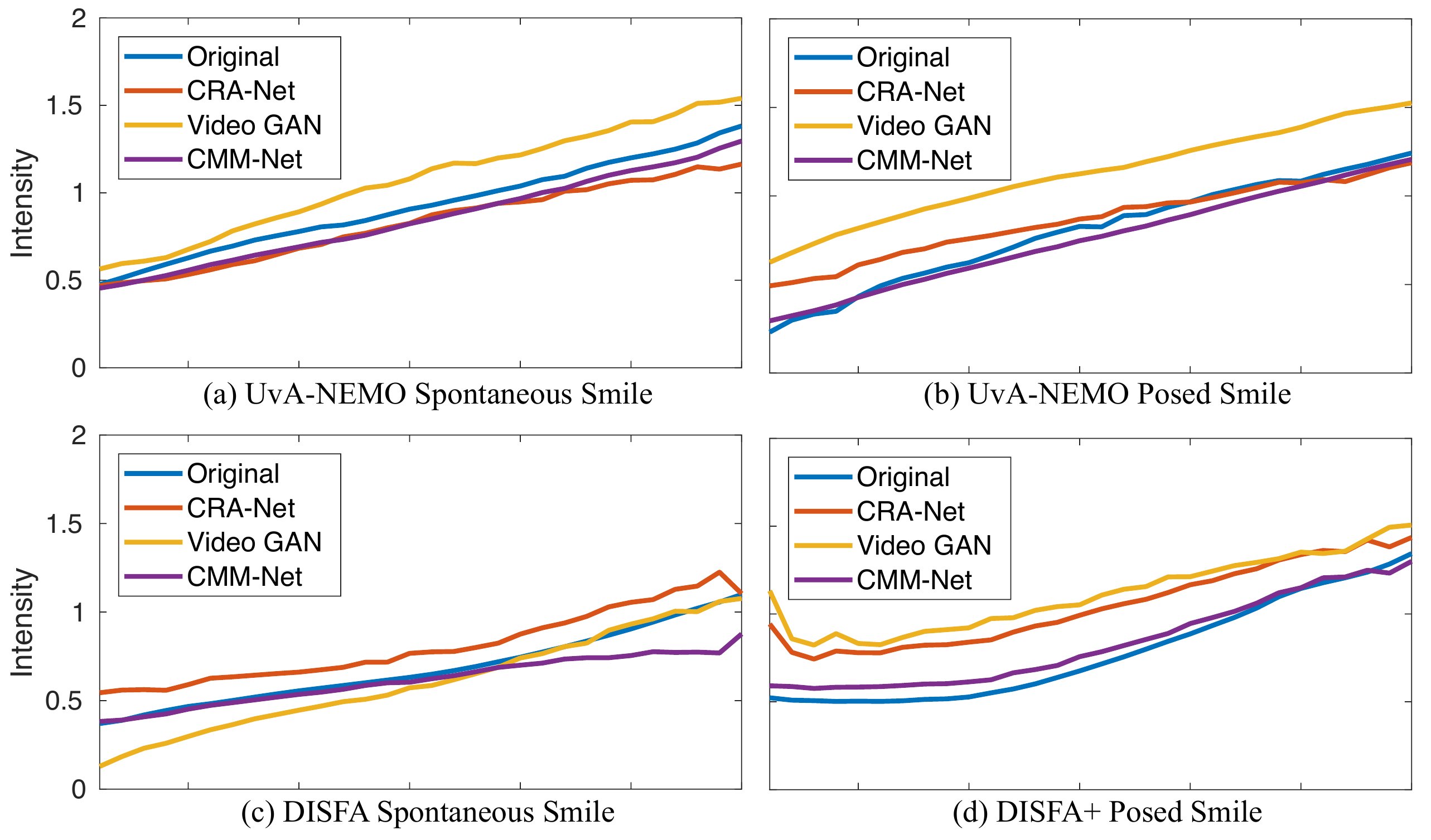}
\vspace{-5mm}
\caption{Dynamics of the action units in N2S sequences.}
\label{fig: au_curves}
\vspace{-5mm}
\end{figure}

\vspace{-5mm}
\paragraph{Analyzing the Dynamics of AUs.} 
In a final series of experiments we evaluate whether the AUs of the generated data have the same dynamics as the original sequences. In detail, we measure the intensity of the \textit{cheek raiser} AU over the generated sequences using the videos from the testing set, smooth it with a 5-frames long window and plot the average over the test set in Fig.~\ref{fig: au_curves}. We clearly observe that the curves closest to the original data are the ones associated to CMM-Net. This demonstrates the advantage of using a landmark image embedding and proves that the multi-mode image sequences have dynamics that are very similar to the real data. For Video-GAN, the generated videos usually have poor quality making it hard to automatically compute the AU score. Thus, the curves of Video-GAN always significantly deviate from the curve corresponding to the original sequence.
Table \ref{tab: AU_distance} shows the cumulative distance between the AU curves of different models and those corresponding to the original sequences. The values reported in the table further confirm the previous observations. 

\begin{table}[t]
\centering
\caption{Distance between the AU curves of different methods and those of the original sequences.\vspace{1mm}}
\label{tab: AU_distance}
\setlength\tabcolsep{3pt}
\resizebox{\columnwidth}{!}{%
\begin{tabular}{ccccc}
\toprule
Model     & UvA-NEMO Spont. & UvA-NEMO Posed & DISFA & DISFA+ \\ \midrule
Video GAN & 2.976       & 2.618      & 3.775 & 7.979  \\ 
CRA-Net   & 4.452       & 9.783      & 2.400 & 9.931  \\ 
CMM-Net   & 2.234       & 1.472      & 2.035 & 1.812  \\ \bottomrule
\end{tabular}}
\vspace{-3mm}	
\end{table}

\vspace{-3mm}
\section{Conclusions}
\vspace{-3mm}
In this paper we address the task of smile generation and, in particular, we show how to synthesize distinct face videos of one person given a facial expression (\eg posed vs. spontaneous smile). We proposed a novel framework which decouples information about facial expression dynamics, encoded into landmarks, and face appearance. For generating landmark sequences we proposed a two layer conditional recurrent network. 
The first layer generates a sequence of facial landmark embeddings conditioned on a given facial expression label and an initial face landmark. The second layer is responsible for generating multiple landmark sequences starting from the output of the first layer.
The landmark sequences are then translated into face videos adopting a U-Net like architecture. The reported experiments on two public datasets demonstrate the effectiveness of our CMM-Net for generating multiple smiling sequences.
In the future, we would like to explore the role of low-level characteristics (e.g.\ attention models or structured multi-scale features~\cite{Xu-NIPS-2017}) of high-level subjective properties~\cite{Alameda-CVPR-2017} in facial expression generation.


\balance
{\small
\bibliographystyle{ieee}
\bibliography{egpaper_final}
}

\end{document}

%% file: related.tex
\section{Related Work}
\label{related}
\vspace{-2mm}
\paragraph{Image Generation.} Recent developments in the deep learning field have brought significant advances in the area of image generation. Deep generative models such as generative adversarial networks (GAN)~\cite{goodfellow2014generative} 
and variational auto-encoders (VAE)~\cite{kulkarni2015deep} have shown to be extremely powerful for synthesizing still images. 

GANs, and in particular conditional GANs \cite{mirza2014conditional}, have been exploited in many applications, \eg to modify the appearance of a picture according to the user inputs \cite{zhu2016generative}, to synthesize faces from landmark images \cite{di2017gp}, 
to translate synthetic images into realistic photos \cite{bousmalis2016unsupervised}, and
for image colorization \cite{isola2016image}. 
Due to the good performance and wide applications, GANs have received an increasing interest lately and several variations over the original model in \cite{goodfellow2014generative} have been introduced, such as CycleGAN~\cite{CycleGAN2017}, DiscoGAN~\cite{kim2017learning}, 
and Wasserstein GAN 
(W-GAN)~\cite{arjovsky2017wasserstein}.
Similarly to GANs, VAEs have also been extensively used to generate images and many VAE-like models have been introduced, such as Gaussian Mixture VAE~\cite{dilokthanakul2016deep}, 
Hierarchical VAE~\cite{goyal2017nonparametric} and VAE-GAN~\cite{larsen2016autoencoding}. VAEs have been exploited for synthesizing images of handwritten digits \cite{salimans2015markov}, pictures of house numbers \cite{gregor2015draw} and future frames \cite{walker2016uncertain}.

Recent works have considered both GANs and VAEs models for generating face images. For instance, in \cite{hou2017deep} a variational autoencoder adopting a perceptual loss is shown to be effective for synthesizing faces while encoding information about facial expressions. In \cite{yan2016attribute2image} the problem of generating face images given some specific attributes, \eg related to age, gender or expressions, is addressed with deep generative models. Similarly, in \cite{li2016deep} a GAN-based model is proposed for transferring facial attributes while preserving as much as possible information about identity. However, these previous works
considered the problem of generating images, while in this paper we explicitly aim to synthesize \textit{face videos} (\eg of smiling people). 


\vspace{-6mm}
\paragraph{Video Generation.} Fostered by the success in image generation, recent works have started to explore deep networks to generate videos \cite{zhou2016learning, vondrick2016generating,saito2016temporal,oh2015action}. 
Two types of approaches have been proposed for this. A first strategy is based on the use of a spatio-temporal network which synthesizes all the frames simultaneously. For instance, in~\cite{vondrick2016generating} a 3D spatio-temporal DCGAN \cite{radford2016unsupervised,ji20133d} is introduced. Similarly, in \cite{saito2016temporal} a temporal generative adversarial network which generates multiple frames of a sequence at the same time is presented. However, these methods usually are typically associated to a poor image quality. %
The second strategy models temporal dependencies by taking advantage of recurrent neural networks (RNNs) which generate images sequentially.
For instance, in \cite{oh2015action} a convolutional long-short term memory (LSTM) network is used to predict the future frames in Atari games conditioned on an action label. In \cite{tulyakov2017mocogan} a gated recurrent neural network (GRU) is employed within an adversarial learning framework to generate videos decoupling appearance from motion information. Similarly, in~\cite{villegas2017learning} the authors proposer a hierarchical prediction pipeline based on LSTMs in order to estimate a sequence of full-body poses and generate a realistic video.


Our work belongs to the second category. However, different from previous studies on face generation, we investigate the use of landmark images which can be extracted using \cite{wang2018recurrentconv,wang2016recurrentconv,baltruvsaitis2016openface} for this purpose. We demonstrate that operating on landmarks we can better encode the dynamics of facial expressions. Furthermore, the landmark manifold space is relatively easier to learn with respect to that associated to the original face images, as landmark images only contain binary values and are very sparse. This  fact has a clear impact in reducing the computational overhead.
The benefits of exploiting landmark information to generate smile sequences are shown in the experimental section. 
To the best of our knowledge, this is the first study proposing a method able to generate multiple sequences given a neutral face image and a conditioning class label. 
Indeed, current video generation models only focus on creating a single sequence and the problem of synthesizing visual contents in a one-to-many setting has only recently been addressed in case of images \cite{ghosh2017multi}.
